\documentclass[runningheads]{llncs}
\usepackage[T1]{fontenc}
\usepackage{amsmath,amssymb,amsfonts}
\usepackage{algorithm}
\usepackage{algpseudocode}
\usepackage{graphicx}
\usepackage{textcomp}
\usepackage{subcaption}
\usepackage{multirow}  
\usepackage{booktabs} 
\usepackage[dvipsnames,table,xcdraw]{xcolor}  % Load xcolor once with all options
\usepackage{multicol}
\usepackage{placeins}  % Loaded once
\usepackage{setspace}
\usepackage{enumitem}
\usepackage{makecell}
\usepackage{caption}
\usepackage{authblk}
\usepackage{orcidlink}

\title{RBLA: Rank-Based-LoRA-Aggregation for Fine-tuning Heterogeneous Models in FLaaS}
\begin{document}

\titlerunning{RBLA}
% If the paper title is too long for the running head, you can set
% an abbreviated paper title here
%
\author{Shuaijun Chen\inst{1}\orcidlink{0009-0001-4944-3406} 
\and Omid Tavallaie\inst{1,2}\orcidlink{0000-0002-3367-1236} 
\and Niousha Nazemi\inst{1}\orcidlink{0000-0003-4085-7044} 
\and Albert Y. Zomaya\inst{1}\orcidlink{0000-0002-3090-1059}}
\authorrunning{S. Chen et al.}
% First names are abbreviated in the running head.
% If there are more than two authors, 'et al.' is used.
%
\institute{School of Computer Science, The University of Sydney, Australia \and Department of Engineering Science, University of Oxford, United Kingdom
\email\{shuaijun.chen, niousha.nazemi, albert.zomaya\}@sydney.edu.au, omid.tavallaie@eng.ox.ac.uk}

\maketitle

%
            % Typeset the header of the contribution
%
\begin{abstract}
Federated Learning (FL) is a promising privacy-aware distributed learning framework that can be deployed on various devices, such as mobile phones, desktops, and devices equipped with CPUs or GPUs. In the context of server-based Federated Learning as a Service (FLaaS), FL enables a central server to coordinate the training process across multiple devices without direct access to local data, thereby enhancing privacy and data security. Low-Rank Adaptation (LoRA) is a method that efficiently fine-tunes models by focusing on a low-dimensional subspace of the model's parameters. This approach significantly reduces computational and memory costs compared to fine-tuning all parameters from scratch. When integrated with FL, particularly in a FLaaS environment, LoRA allows for flexible and efficient deployment across diverse hardware with varying computational capabilities by adjusting the local model's rank. However, in LoRA-enabled FL, different clients may train models with varying ranks, which poses challenges for model aggregation on the server. Current methods for aggregating models of different ranks involve padding weights to a uniform shape, which can degrade the global model's performance. To address this issue, we propose Rank-Based LoRA Aggregation (RBLA), a novel model aggregation method designed for heterogeneous LoRA structures. RBLA preserves key features across models with different ranks. This paper analyzes the issues with current padding methods used to reshape models for aggregation in a FLaaS environment. Then, we introduce RBLA, a rank-based aggregation method that maintains both low-rank and high-rank features. Finally, we demonstrate the effectiveness of RBLA through comparative experiments with state-of-the-art methods.

\end{abstract}

\section{Introduction}
\begin{figure}[t]
\centering
\begin{subfigure}{0.45\textwidth}
\includegraphics[width=0.9\linewidth]{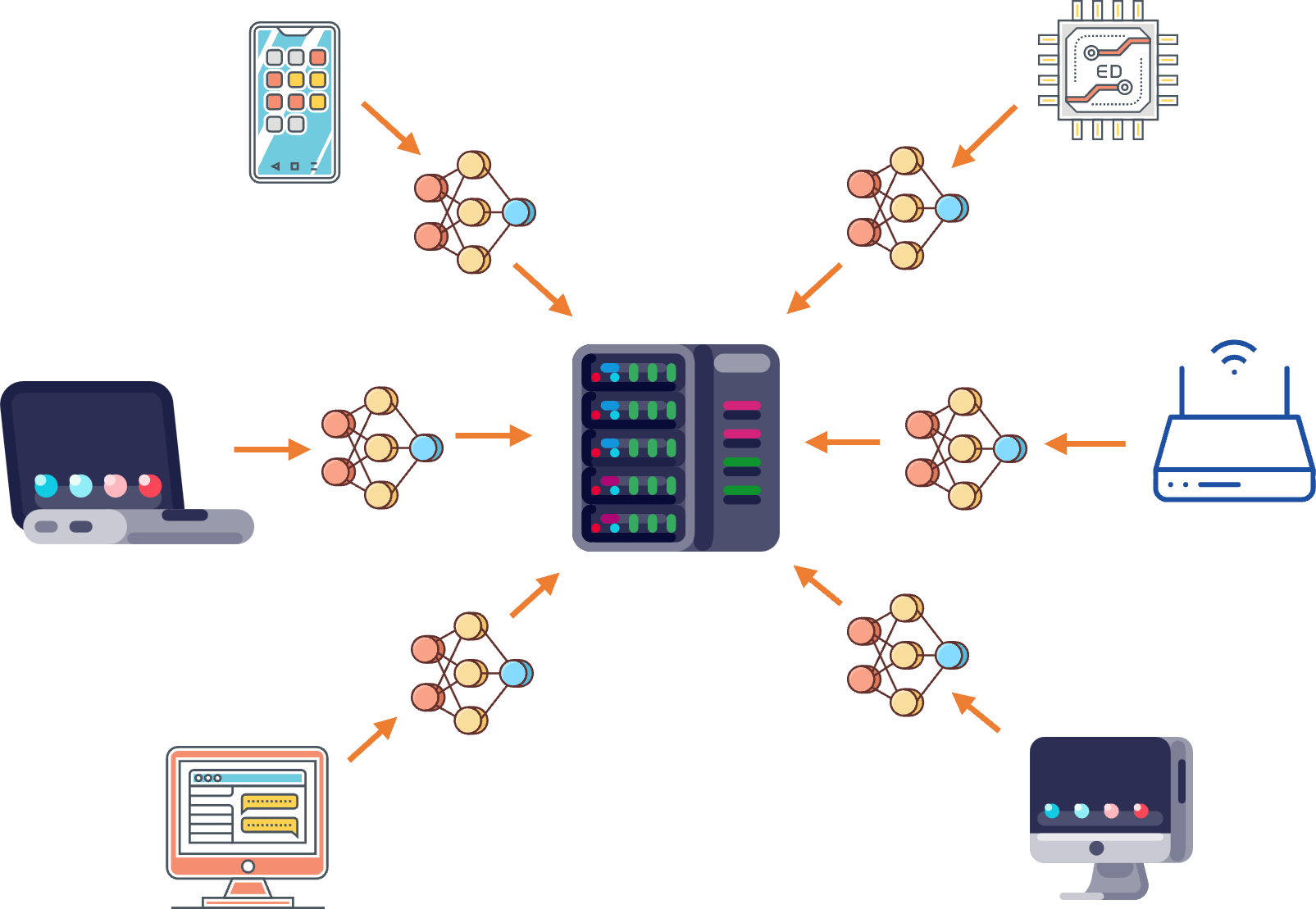} 
\caption{FlaaS with varied devices}
\label{fig:fl}
\end{subfigure}
\begin{subfigure}{0.45\textwidth}
\centering
\includegraphics[width=0.7\linewidth]{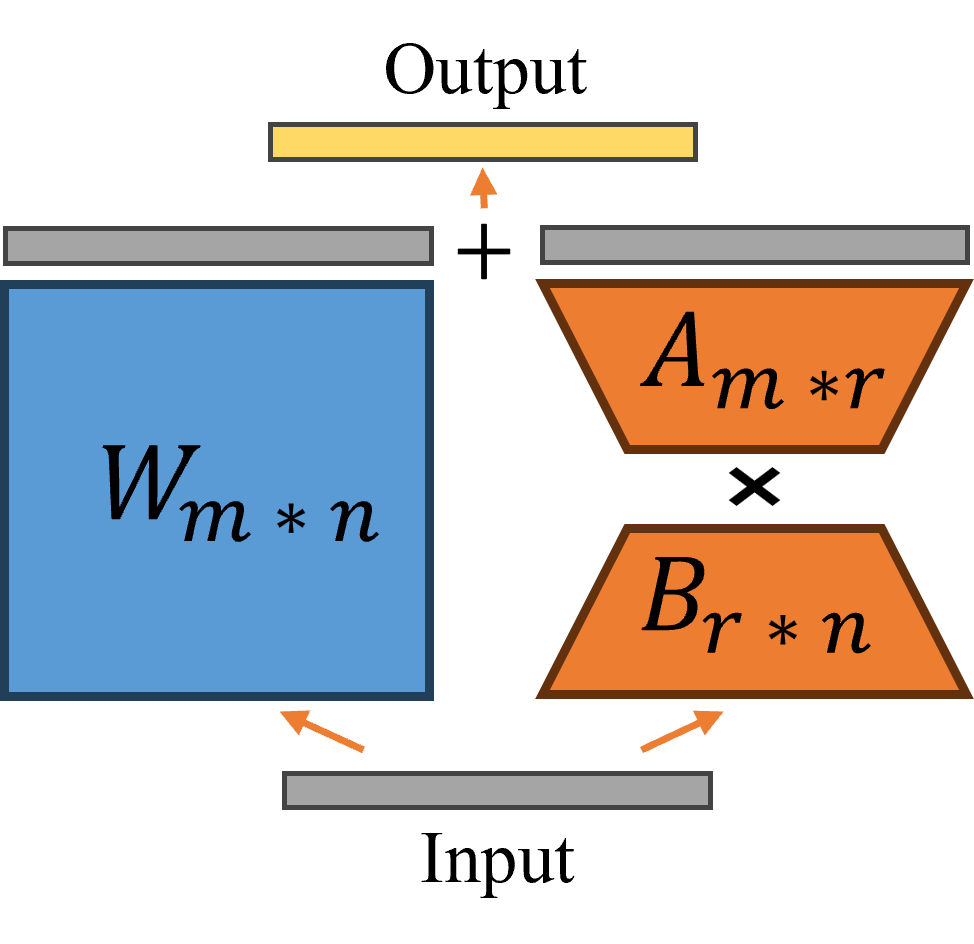}
\caption{Low-Rank Adaptation (LoRA)}
\label{fig:lora}
\end{subfigure}
\caption{a) An application for FLaaS where devices with heterogeneous computational resources (such as smartphones, laptops, and routers) train models with different architectures and send them to a central server for aggregation (without sharing raw data). b) Reducing the dimension of the trained model in the LoRA technique by decomposing the original weight matrix into two smaller matrices to optimize computational efficiency.}\vspace{-5mm}
\label{fig:fl_and_lora}
\end{figure}

Neural Networks (NN) have become a widely applied approach in contemporary Computer Vision (CV) and Natural Language Processing (NLP). Traditionally, model training involves collecting task-specific data and conducting centralized training in high-performance data centers. However, this centralized method raises significant privacy concerns, especially in applications dealing with sensitive data such as health information. Collecting and uploading user data to centralized servers can inherently lead to privacy breaches. To address these privacy challenges, McMahan et al. proposed Federated Learning (FL) in 2017 \cite{mcmahan2017communication}. FL is a privacy-preserving distributed machine learning framework designed for decentralized neural network training \cite{nazemi2024access}. This algorithm trains models locally on client devices (without sharing raw data), and model updates are then uploaded to a central server for aggregation. This concept has further evolved into Federated Learning as a Service (FLaaS) \cite{flaas,gao2024federated}, shown in Figure .\ref{fig:fl}, where federated learning capabilities are provided as a cloud-based service \cite{chen2023boosting}. FLaaS simplifies the deployment process by managing the underlying infrastructure, allowing the implementation of federated learning across a diverse range of devices, including smartphones, IoT devices, and edge servers. This flexibility enables organizations to leverage privacy-preserving machine learning on a large scale without the complexities of building and maintaining their own FL systems. Despite these advantages, decentralized local model training in FL typically requires robust client device performance and communication capabilities. In practical FlaaS scenarios, significant performance variations among client devices pose challenges in deploying models with identical structures across all clients. 

Recent advancements in computing power and algorithms have led to numerous applications that use large models to process vast amounts of data on mobile devices, providing responses or making decisions. Examples include ChatGPT and Tesla's autonomous driving technology. However, the inference and training processes of large models are complex and resource-intensive. Additionally, the heterogeneous nature of client devices in FL scenarios, with significant differences in performance, complicates the deployment and fine-tuning of federated large models. To address these issues, Low-Rank Adaptation (LoRA) \cite{hu2022lora} has emerged as a feasible solution. Figure .\ref{fig:lora} shows LoRA decomposes the locally deployed model into two low-rank matrices, which can then be adjusted in matrices' ranks based on the local data or device performance demands. This approach decreases both the model training cost and performance requirements and allows each client to customize the model size to its computational capacity and data characteristics. As a result, LoRA makes training and deployment more efficient.

\begin{figure}[t]
  \centering
    \includegraphics[width=0.85\linewidth]{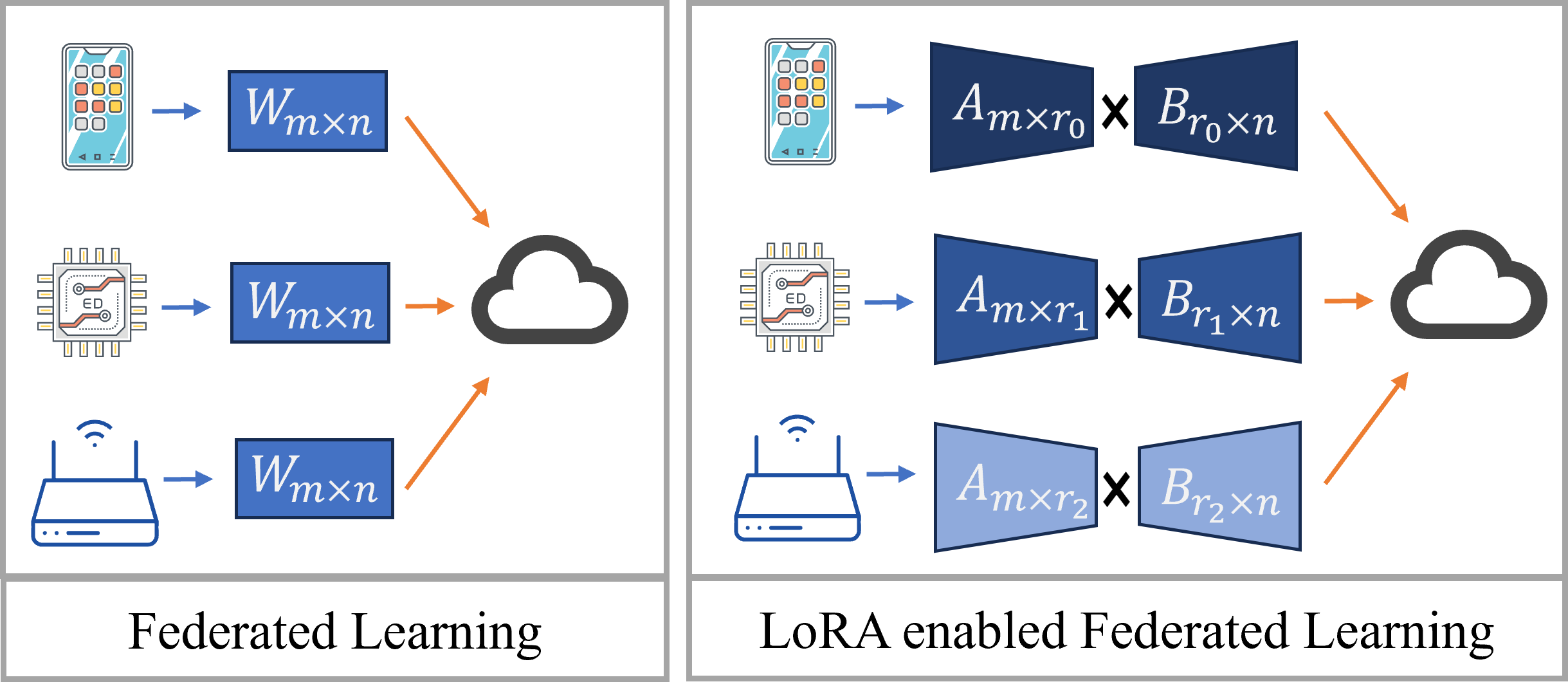}
  \caption{Comparison between traditional FL and LoRA-enabled FL. The left figure shows the standard FL, where clients have full-rank \(m \times n\) weight matrices. In contrast, the right figure illustrates LoRA-enabled FL, where clients send low-rank matrices (trained models) with varying ranks to reduce communication and computation costs.}\vspace{-5 mm}
  \label{fig:methods_comparision}
\end{figure}

Figure. \ref{fig:methods_comparision} illustrates the differences between traditional centralized training, FedAvg, and FL with LoRA. As shown in the figure, unlike traditional federated learning, FL with LoRA allows each client to customize the rank number based on its own conditions \cite{cho2023heterogeneous}. This flexibility enables the system to adapt to the diverse needs and capabilities of different clients. However, this customization introduces a challenge during the server aggregation process: matrices of different ranks cannot be directly aggregated using traditional methods. To address this, all matrices must be expanded to a uniform dimension for computation.

The current matrix expansion method primarily relies on zero-padding \cite{cho2023heterogeneous}, a technique that involves padding smaller matrices with zero elements to match the dimensions of the largest matrix. While this method is straightforward to implement, it introduces significant issues. First, zero-padded matrices contain a large number of zero elements, leading to unnecessary computational overhead during the aggregation process. Second, padding with zeros can degrade model performance, as the padded zeros do not provide useful information during aggregation. Additionally, the dimensional expansion of matrices can result in a substantial increase in memory usage, which is particularly problematic for resource-constrained devices.

We propose Rank-Based Layer Aggregation (RBLA), a method designed to effectively aggregate matrices of varying ranks to address the issues posed by heterogeneous matrix ranks in FL. RBLA aims to preserve the high-dimensional features generated during the training process, mitigate the negative impacts of matrix aggregation, and enhance the global model's convergence speed. The main contributions of our work are as follows:

\begin{enumerate}[topsep=0 pt, partopsep=0 pt, wide=0 pt]
\item We present a toy example demonstrating weight aggregation in LoRA-enabled FL and explain the effects of zero-padding on higher-dimensional features. 

\item We propose RBLA, a model aggregation method specifically designed to aggregate LoRA models of different ranks.

\item We provide a Python implementation of RBLA using the TensorFlow library and compare its performance against state-of-the-art methods across multiple datasets.
\end{enumerate}

\section{Related Work}
FL is a decentralized model training method deployable on any device with computational resources and internet connectivity. In practical FL scenarios, clients often possess distinctly different local datasets \cite{9835537}, which may not accurately represent the overall data distribution. Additionally, client device performance is usually heterogeneous \cite{deng2020edge} due to variations in computational power, memory, and network bandwidth \cite{liu2022distributed}. This heterogeneity can lead to challenges in effectively training and aggregating the global model \cite{hsieh2020non,nazemi2024boosting}, as some clients may contribute less due to their limited resources \cite{li2020federated}. The non-IID nature of data can result in a global model that may not generalize effectively across all clients \cite{zhao2018federated}. 

Various methods have been proposed to address the data non-IID problem. Collins et al. \cite{collins2021exploiting} proposed FedRep, which learns a shared representation across clients while allowing each to train personalized local models. Similarly, Ghosh et al. \cite{ghosh2020efficient} proposed the IFCA, which clusters clients with similar data distributions and optimizes models within each cluster to reduce the impact of data heterogeneity. In the context of personalization, Arivazhagan et al.\cite{arivazhagan2019federated} introduced FedPer, which splits models into shared and personalized layers. FedAvg is used to train shared layers and local-update-personalized layers to obtain better end-side performance. Furthermore, to tackle client drift caused by data heterogeneity, Karimireddy et al. \cite{karimireddy2020scaffold} proposed SCAFFOLD, an algorithm that uses control variates to correct client drift and achieve quicker convergence. To address the issue of device heterogeneity, Khodadadian et al. \cite{khodadadian2022federated} tackled client communication costs by utilizing TD-learning and Q-learning to reduce communication overhead. Kumar et al. 
 \cite{kumar2023federated} applied LoRaWAN technology to FL to improve communication efficiency and robustness. Li et al. \cite{li2020federated} proposed FedProx, designed to tolerate not-completely-trained local models and use a proximal term to reduce the impact of over-fitted local models. In practical FlaaS involving multiple IoT devices, the expected local model architecture may differ among client devices due to heterogeneous hardware configurations, tasks, and personal demands \cite{challenge1}, \cite{challenge}, and \cite{challenge2}. To perform effective aggregation on a central server with models of different architectures, Wang et al. \cite{wang2023flexifed} proposed MaxCommon which facilitates collaborative training across various models. To further enhance the flexibility and efficiency of FL in heterogeneous environments, LoRA \cite{hu2022lora} has emerged as a promising technique. LoRA allows flexible adaptation of trainable parameters based on data quality, device performance, and other factors. Based on LoRA and FL, Yi et al. \cite{yi2023fedlora} proposed pFedLoRA, which incorporates LoRA in FL to increase model fine-tuning efficiency. Choi et al. proposed HetLoRA \cite{cho2023heterogeneous}, which dynamically adjusts the client's rank based on the local model's quality of fit to the local data. Other approaches, such as DFLNet \cite{zhang2023dflnet}, aim to propose to improve secure authentication and model convergence in LoRA-enabled networks. LLDPC \cite{yang2024low} enhances data transmission reliability in LoRA networks. FDLoRA \cite{qi2024fdlora} balances personalized and global learning for large language models (LLMs) while reducing costs, and DP-LoRA \cite{liu2023differentially} ensures differential privacy in FL for LLMs with minimal communication overhead.

\section{Problem Statement}
% \begin{table}[b]
%     \caption{Declaration of notations}
%     \centering
%     \begin{tabular}{|c|c|}
%     \hline
%     Notation & Definition \\
%     \hline
%     $i$ & Client $i$\\
%     \hline
%     $W_i$ & Weight of client $i$\\
%     \hline
%     $C_{m \times n}$ & Weight matrix $C$ with $m$ rows and $n$ columns \\
%     \hline
%     $C_r$ & row $r$ of weight matrix $C$\\
%     \hline
%     $w_i$ & Aggregation weight of client $i$\\
%     \hline
%     $A_{p \times q}$ & Weight matrix $A$ with $p$ rows and $q$ columns\\
%     \hline
%     $A'$ & Padded weight matrix of $A$\\
%     \hline
%     $\delta_{i,r}$ & Indicator function result of client $i$ in row $r$\\
%     \hline
%     $E$ & Number of epochs\\
%     \hline
%     $B$ & Local minibatch size\\
%     \hline
%     \end{tabular}\vspace{3 mm}
%     \label{table:declaration_of_notations}
% \vspace{-8 mm}\end{table}

In FL, models must have the same shape to be aggregated on the central server, and zero-padding is one of the most commonly used methods to match dimensions. However, we found that zero-padding introduces structural sparsity, which can slow down convergence. This section presents the impact of zero-padding of trained model weights on the aggregated global model. We specifically analyze how zero-padding affects the sparsity and effectiveness of weighted averaging in the aggregation of LoRA weights \cite{cho2023heterogeneous}. Consider two weight matrices $A_{p \times q}$, $B_{m \times n}$ ($pq<mn$) with corresponding aggregation weights $w_1$ and $w_2$. To aggregate models with different dimensions, matrix $A$ is padded with zeros to match the dimensions of $B$, resulting in the padded matrix $A'$ with a shape of $m \times n$. The aggregated model $C_{m \times n}$ is then computed as:
\begin{equation}
C = w_{1}A' + w_{2}B.
\end{equation}
Here, the term 'zero-padding' introduces non-informative values that negatively impact the overall feature representation, impairing the neuron network's ability to effectively generalize target features \cite{neyshabur2017exploring}. We highlight the significance of structural sparsity and its impact on neural network capacity and learning dynamics (\cite{neyshabur2017exploring}, \cite{zhang2021understanding}, \cite{goodfellow2016deep}, and \cite{kawaguchi2017generalization}), which underscores the need to avoid bringing structural sparsity during deep feature extraction when aggregating LoRA weights across different dimensions. For example, during the LoRA training process, model $A$ may fail to capture the depth features learned by model $B$ due to insufficient neurons. During aggregation, zero-padding causes the layers of matrix $A$ that are padded with zeros to dilute the depth features trained by matrix $B$, as these padded layers cannot provide the necessary non-linear transformations \cite{neyshabur2017exploring}. Consequently, zero-padding introduces invalid zero-value information, which dilutes the important depth features learned by model $B$ during the averaging process. This scenario can be illustrated as follows:

\begin{equation}
A' = \begin{bmatrix}
a_{11} & a_{12} & a_{13} \\
a_{21} & a_{22} & a_{23} \\
0 & 0 & 0
\end{bmatrix}, \quad
B = \begin{bmatrix}
b_{11} & b_{12} & b_{13} \\
b_{21} & b_{22} & b_{23} \\
b_{31} & b_{32} & b_{33}
\end{bmatrix}, \quad
C = \begin{bmatrix}
c_{11} & c_{12} & c_{13} \\
c_{21} & c_{22} & c_{23} \\
c_{31} & c_{32} & c_{33}
\end{bmatrix}.
\end{equation}
The last row of matrix $C$ by using weighted average can be represented as:

\begin{equation}
  C_{3,n} = \frac{w_1 \cdot a'_{3,n}}{w_1 + w_2} + \frac{w_2 \cdot b_{3,n}}{w_1 + w_2} = \frac{w_2 \cdot b_{3,n}}{w_1 + w_2}, \quad a'_{3,n} = \mathit{0}_{1 \times 3}.
  \label{eq:last_row}
\end{equation}
Eq. \ref{eq:last_row} presents how zero-padding results in a loss of information by diluting the features captured by the deeper model during aggregation. Extending this scenario to $n$ weight matrices $A_1, A_2, \ldots, A_n$ with varying dimensions, let the largest matrix be of size $m \times n$. For simplicity, assume the aggregation weights are $w_1, w_2, \ldots, w_n$. For each matrix $A_i$ with dimensions $p_i \times q_i$ where $p_i q_i < mn$, zero-padding is applied to create $A_i'$, a matrix of dimensions $m \times n$.

\begin{equation}
    A_i' = 
    \begin{cases} 
    A_{kj}, & 1 \leq k \leq p \text{ and } 1 \leq j \leq q ,\\
    0, & \text{otherwise}.
    \end{cases}
    \label{eq:zero_padding}
\end{equation}
For each matrix $A_i$ with dimensions $p_i \times q_i$ where $p_i q_i < mn$, we apply zero-padding to create $A_i'$ with dimensions $m \times n$ by Eq. \ref{eq:zero_padding}. The row $C_r$ 
of the aggregated global model $C$ is computed using a weighted average, as below:

\begin{equation}
    C_r = \frac{\sum_{i=1}^{n} w_i \cdot \delta_{i,r} \cdot \mathbf{a}_{i,r}}{\sum_{i=1}^{n} w_i}, \delta_{i,r} = 
    \begin{cases} 
    1, & \text{if } $r < n$, \\
    0, & \text{otherwise.}
    \end{cases}  
    \label{eq:zero_padding_with_n_matrix}
\end{equation}
Here, $\mathbf{a}_{i,r}$ represents the $r$-th row vector of the zero-padded matrix $A_i'$, and $\delta_{i,r}$ is an indicator function that equals 1 if the $r$-th row of $A_i$ exists and 0 otherwise. This indicates that \textbf{the more zero-padded is applied to a layer, the more the original features of that layer become diluted}. Consequently, zero-padding introduces significant issues as it incorporates \textbf{a substantial number of zero values}, which dilute effective feature information during aggregation. This dilution reduces the impact of high-dimensional features learned by models with higher dimensions. These zero values, acting as invalid information in the computation, degrade the quality of the aggregated results and lead to a model populated with numerous irrelevant features. In scenarios with skewed data distributions, such as long-tailed distributions where low-rank model clients are assigned fewer classes or data, high-dimensional features learned by deeper neural networks are further diluted by shallower networks during zero-padding weighted average aggregation. This results in additional performance degradation, as zero-padding often diminishes the features from client models with smaller datasets. This issue prevents the aggregated model from fully leveraging all clients' data and features. Moreover, the \textbf{structural sparsity} introduced by zero-padding imposes invalid linear transformations, which limit the model's capacity to represent complex patterns and reduce its effectiveness in learning high-dimensional spaces. Additionally, structural sparsity restricts the model's generalization capability and lowers the overall performance. These factors collectively lead to a decline in accuracy and robustness, significantly impacting the model's effectiveness in practical applications.

\section{RBLA}

\begin{algorithm}[t]
\caption{Server Aggregation of RBLA. The server aggregates weights from all clients by accurately handling shared and unique layers using an indicator function and weighted aggregation.}
\label{alg:server}
\begin{algorithmic}
    \State Initialize $W_{\text{agg}} \gets 0$

    \State Receive all $W_i$ from client $i$

    \For{each layer $W_{\text{agg}, r}$}
        \State Initialize $w_r \gets 0$
        \For{each model $i = 1, 2, \ldots, N$}
            \If{$\delta_{i, r} = 1$}
                \State $W_{\text{agg}, r} = W_{\text{agg}, r} + w_i \cdot W_{i,r}$
                \State $w_r = w_r + w_i$
            \EndIf
        \EndFor
        \State $W_{\text{agg}}$.append($W_{\text{agg}, r} / w_r$)
    \EndFor
    
    \State $W_{\text{server}} = W_{\text{agg}}$
    \State Send $W_{\text{server}}$ to all clients

\end{algorithmic}
\end{algorithm}

\begin{figure}[t]
  \centering
    \includegraphics[width=0.9\linewidth]{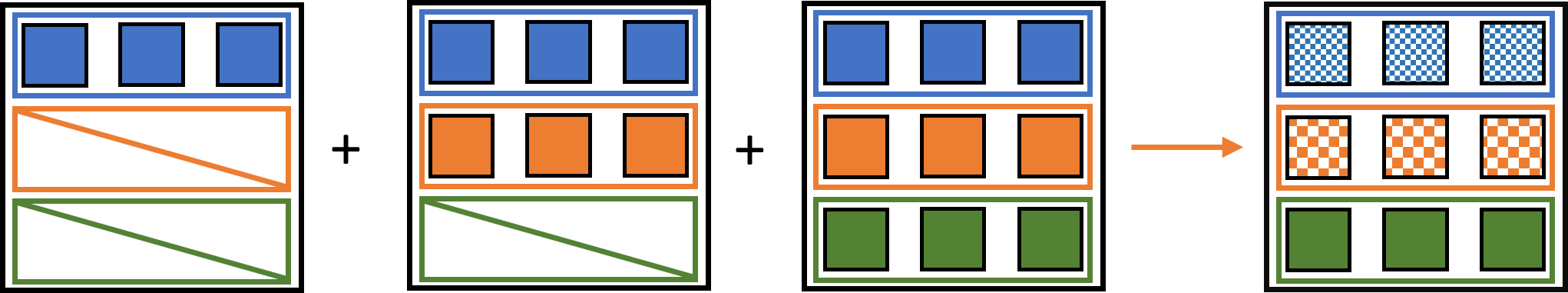}
  \caption{Horizontal aggregation process for heterogeneous LoRA models: Similar to vertical aggregation, RBLA preserves the weights of shared layers and retains unique layers during the horizontal aggregation process.}\vspace{-1 mm}
  \label{fig:horizontal_aggregation}
\end{figure}

\begin{figure}[t]
  \centering
    \includegraphics[width=0.9\linewidth]{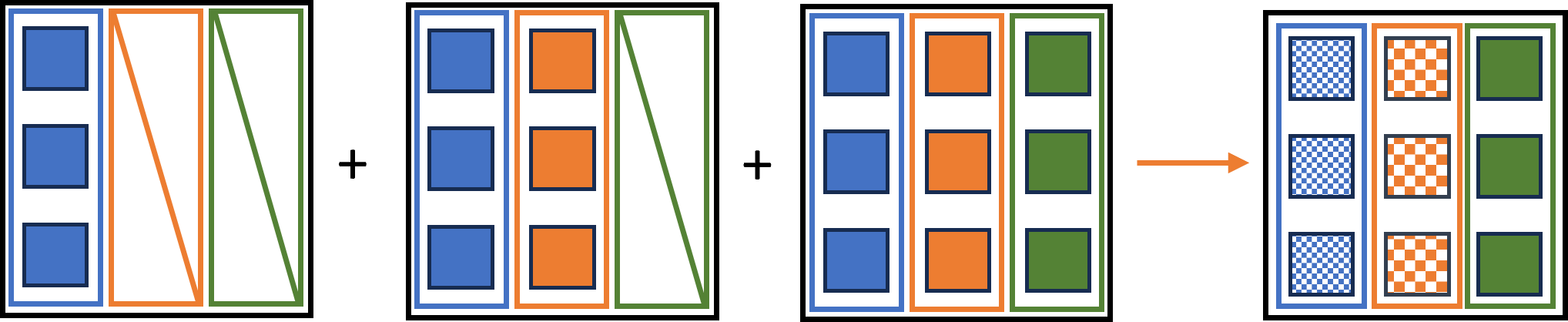}
  \caption{Vertical aggregation process for heterogeneous LoRA models. Layers of the same color represent common layers. RBLA aggregation preserves the original value of unique layers and calculates a weighted average for common layers.}\vspace{-1 mm}
  \label{fig:vertical_aggregation}
\end{figure}

RBLA is designed to aggregate heterogeneous model weight matrices, bias matrices, and low-rank matrices of different ranks from multiple clients by re-weighting the aggregation weights. Algorithm. \ref{alg:server} and Algorithm. \ref{alg:client} show the procedures executed on the server and client, respectively. Considering matrix sparsity, RBLA first pads all matrices to match the dimensions of the largest LoRA matrix, filling missing entries with a neutral value. It then performs a weighted aggregation for the \textbf{existing common layer values}, either row-wise or column-wise, depending on the presence of none-values and their corresponding aggregation weights. During this procedure, RBLA calculates the aggregation weights based on the common layers, performs a weighted average for shared layers, and preserves the original value of unique stand-alone layers. Figures \ref{fig:vertical_aggregation} and \ref{fig:horizontal_aggregation} illustrate the RBLA aggregation process for models with heterogeneous columns and rows, respectively.

\begin{algorithm}[b]
\caption{The client procedure in RBLA.}
\label{alg:client}
\begin{algorithmic}
    \State \textbf{Clients receive $W_{server}$}
    \State $p, q \gets \text{shape of } W_{i}$
    \State Extract the $p \times q$ sub-matrix from $W_{server}$
    \State $W_{i} = W_{server}[0:p, 0:q]$
    \For{$\forall e \in E$}
        \For{$\forall b \in B$}
            \State $W_{i} = W_{i} - \eta \nabla \ell(w; b)$
        \EndFor
    \EndFor
    \State Send $W_{i}$ to the server    
\end{algorithmic}
\end{algorithm}

The detailed aggregation process of RBLA is as follows: suppose there are $N$ models indexed by $i = 1, 2, \ldots, N$, each with a weight matrix $W_i$, representing a LoRA model trained at different ranks. To aggregate these models, we use a weighted average to combine shared layers that exist across multiple weight matrices with similar structures. We also preserve unique layers that are present only in a specific matrix. To identify shared layers and unique layers, we define an indicator function $\delta_{i, r}$ as follows:

\begin{equation}
    \delta_{i, r} = 
    \begin{cases} 
    1, & \text{if matrix } \mathbf{A}_i \text{ contains the } r \text{-th layer,} \\
    0, & \text{otherwise.}
\end{cases}
\end{equation}
The aggregated weight for the $r$-th layer, $\mathbf{C}_{r}$, is computed as:
\begin{equation}
    \mathbf{C}_r = \frac{\sum_{i=1}^{n} \delta_{i, r} \cdot w_i \cdot \mathbf{A}_{i,r}}{\sum_{i=1}^{n} \delta_{i, r} \cdot w_i}.
    \label{eq:rbla_main}
\end{equation}
In this process:
\begin{itemize}
    \item \(\mathbf{A}_{i,r}\) represents the \(r\)-th layer of the weight matrix for model \(i\).
    \item \(w_i\) is the weight coefficient for model \(i\).
    \item The indicator function \(\delta_{i, r}\) ensures that only the matrices containing the \(r\)-th layer can contribute to the aggregation.
\end{itemize}
Applying this method, we can aggregate the weights of shared layers while preserving the unique layers of each model to avoid unnecessary structural sparsity.

\section{Experiments and Evaluation}
In our study, we evaluate the effectiveness of RBLA using MLP and CNN architectures on the MNIST, FMNIST, CIFAR-10, and CINIC-10 \cite{darlow2018cinic} datasets, comparing its performance against two baselines: Zero-Padding (ZP) \cite{cho2023heterogeneous} and Full Fine-Tune (FFT) FedAvg \cite{mcmahan2017communication}. The models are implemented using the TensorFlow library in Python, with three neural network models included. Additionally, all experiments are conducted with a \textbf{fixed seed of 42} to ensure reproducibility. All experiments within the same subfigure share the same configuration, differing only in the aggregation algorithms used.

\subsection{Experiments setup}
In the experiment, the MNIST MLP model consists of two hidden layers with 200 neurons each, activated by ReLU, and a 10-class softmax output layer suitable for flattened 28x28 pixel images (784-dimensional vectors). The MNIST CNN model comprises two convolutional layers with 32 and 64 filters, followed by MaxPooling layers, a 512-unit fully connected layer, and a 10-class softmax output layer designed for 28x28 inputs. Both models were trained with a batch size of 64 using the SGD optimizer with a learning rate of 0.01. For the CIFAR dataset, we used a model that includes three convolutional layers: batch normalization, max pooling, dropout, and fully connected layers. The first two sets of convolutional layers have 32 and 64 filters with a 3x3 kernel size and ReLU activation, followed by pooling layers for downsampling and dropout layers for regularization. The feature maps are then flattened and passed through two fully connected layers with 512 neurons each, followed by dropout, and finally output through a 10-class softmax layer. The model that we used for the CINIC dataset has two extra dense layers with 512 neurons each compared to the CIFAR model. The optimizer for the CIFAR and CINIC experiments was set to Adam \cite{kingma2015adam}, and LoRA is applied only to dense layers for all experiments. Additionally, we tested two participation scenarios: one where all clients participate in each training round and another where 20\% of clients are randomly selected to participate in each communication round.

\subsection{Non-iid \& Model rank settings}
In our experiments, data is allocated to each client following two "staircase" patterns to simulate a realistic data distribution, where the label distribution within each client exhibits a long-tail "stair" pattern. These patterns reflect real-world scenarios, such as medical systems, where data from different hospitals or clinics may have \textbf{varying levels of data complexity or diversity}. For instance, some specialized hospitals might focus on specific types of diseases, while general hospitals handle a broader range of disease categories with more patients. Likewise, in sensing device systems, different devices may collect varying types and amounts of data. For example, embedded temperature control devices primarily gather temperature-related data, whereas smart devices (such as smartphones) may involve a wider variety of data, including GPS, step count, sensor data, etc. Based on this, the label distribution in each client's data in our experiment follows a long-tail distribution. Each subsequent client has an increasing number of labels with non-zero sample counts, starting from Client 1, which has samples only for Label 0. As more clients are added, they progressively add more labels, culminating in Client 10, which has a large number of samples for all labels from 0 to 9. 
Simultaneously, the rank ratio of the LoRA model assigned to each client is scaled based on the number of labels each client possesses, with the rank ratio increasing by 0.1 for each additional label. This approach ensures that clients with more labels are allocated higher ranks to better capture the complexity of their data, while those with fewer labels receive lower ranks.

\subsection{Evaluation Results}

\begin{table*}[t]
\centering
\vspace{-4mm}
\caption{The minimum number of training rounds used for each method to achieve the target test accuracy for the global model in full participation experiments. N/A indicates that the corresponding aggregation strategy cannot achieve the target accuracy within 50 training rounds. All experiments in the same column are conducted under the same settings.}
\resizebox{0.9\textwidth}{!}{%
\begin{tabular}{@{}ccccccc@{}}
\toprule
& \multicolumn{2}{c}{MNIST} & \multicolumn{2}{c}{FMNIST} & CIFAR & CINIC\\
\cmidrule(lr){2-3} \cmidrule(lr){4-5}
Method & MLP (95\%) & CNN (98\%) & MLP (83\%) & CNN (98\%) & CNN (48\%) & CNN (40\%) \\
\midrule
ZP & N/A (94.87\%) & 11 (98\%) & N/A (82.87\%) & 24 (97.04\%) & 22 (48.73\%) & 24 (40\%)\\
FFT & 40 (95.04\%) & 22 (98.03\%) & 19 (83.04\%) & N/A (91.36\%) & \textbf{9} (49.16\%) & 14 (40.09\%) \\
\midrule
RBLA & \textbf{11} (95.06\%) & \textbf{4} (98.27\%) & \textbf{7} (83.15\%) & \textbf{7} (98.09\%) & 12 (48.12\%) & \textbf{12} (40.03\%)\\
\bottomrule
\end{tabular}%
}\vspace{-5mm}
\label{tab:result}
\end{table*}

\begin{figure}[t]
\centering
\begin{subfigure}{0.4\textwidth}
\includegraphics[width=0.9\linewidth]{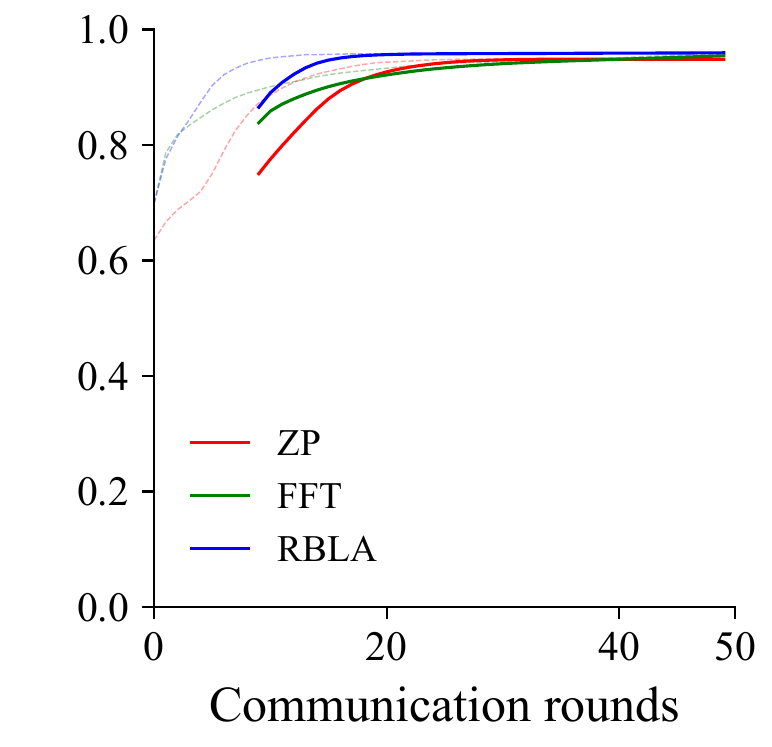}
\vspace{-1mm}
\caption{Full participation}
\label{fig:mnist_all_participate_2nn}
\end{subfigure}
\begin{subfigure}{0.4\textwidth}
\includegraphics[width=0.9\linewidth]{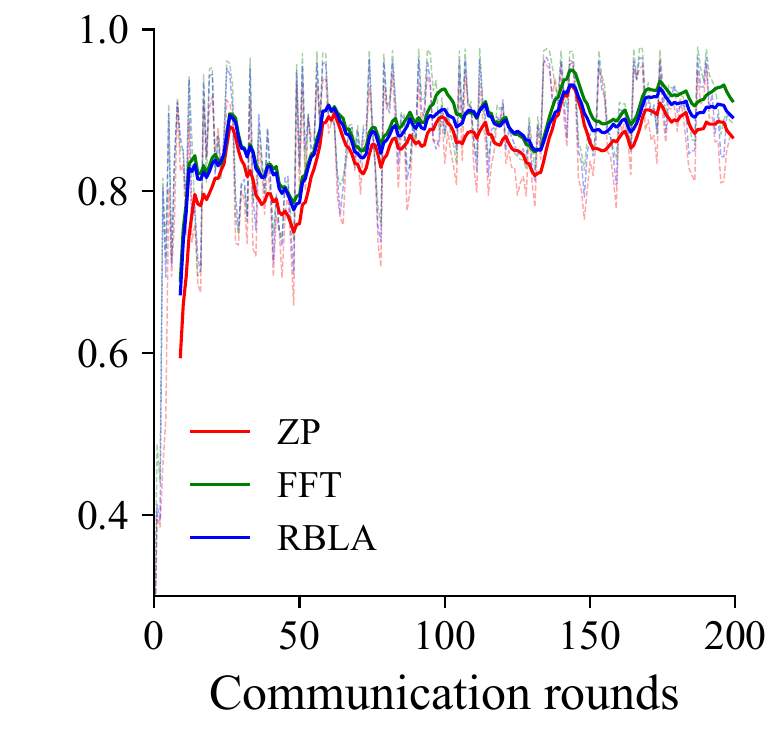}
\vspace{-1mm}
\caption{Random selection}
\label{fig:mnist_random_selection_2nn}
\end{subfigure}
\vspace{-1mm}
\caption{Evaluation of RBLA for \textbf{MNIST} dataset with MLP model.}
\label{fig:mnist_2nn}
\end{figure}

\begin{figure}[t]
\centering
\begin{subfigure}{0.4\textwidth}
\includegraphics[width=0.9\linewidth]{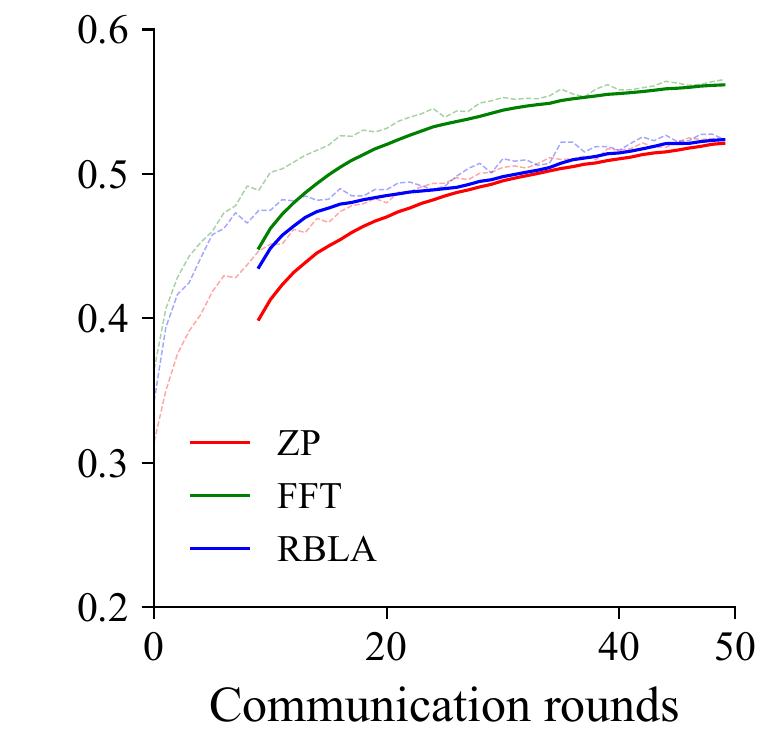}
\vspace{-1mm}
\caption{Full participation}
\label{fig:cifar_all_participate_cnn}
\end{subfigure}
\begin{subfigure}{0.4\textwidth}
\includegraphics[width=0.9\linewidth]{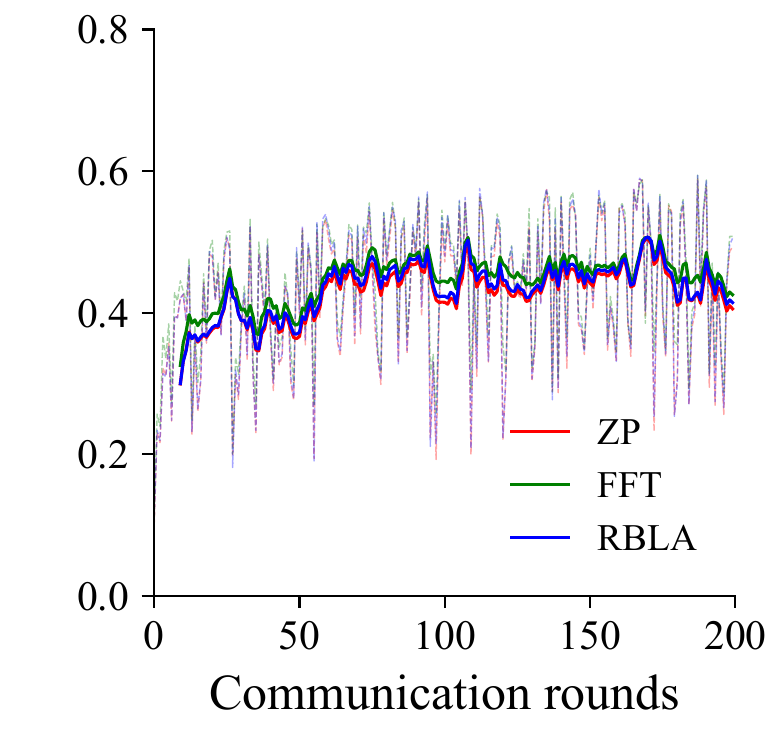}
\vspace{-1mm}
\caption{Random selection}
\label{fig:cifar_random_selection_cnn}
\end{subfigure}
\vspace{-1mm}
\caption{
Evaluation of RBLA for \textbf{CIFAR-10} dataset with CNN model.}\vspace{-3mm}
\label{fig:cifar_cnn}
\end{figure}

In this section, we present the evaluation results across various datasets with different configurations. Table. \ref{tab:result} shows the exact number of communication rounds taken for each method to reach the target test accuracy of the global model. The target accuracy was selected based on the global model accuracy at the communication round where RBLA's convergence speed significantly decreases, and this accuracy can also be reached or nearly reached by the other two methods. Figure. \ref{fig:mnist_2nn} to Figure. \ref{fig:cifar_cnn} show the learning curve of the global model's test accuracy to training rounds; the left sub-figure shows the result with full participants, and the right sub-figure shows the result with random client selection. As an example, the target test accuracy set for the MNIST MLP experiment is \textbf{95\%}, where the convergence of RBLA significantly reduced. The maximum test accuracy of ZP can reach during training is \textbf{94.87\%} (presented in the parentheses), while FFT reaches \textbf{95.04\%} test accuracy in the \textbf{40th} training round, and RBLA reaches \textbf{95.06\%} in the \textbf{11th} training round. \textbf{To ensure clear visual comparisons, a rolling average with a window size of 10 was applied to smooth the data}, represented by solid lines, and the dotted lines illustrate the original and unsmoothed results. The effectiveness of RBLA is assessed on the MNIST, FMNIST, and CIFAR-10 datasets using MLP and CNN models under both full participation and random selection settings. These experiments compare the performance of RBLA against zero-padding and full model fine-tuning.

\begin{figure}[t]
\centering
\begin{subfigure}{0.4\textwidth}
\includegraphics[width=0.9\linewidth]{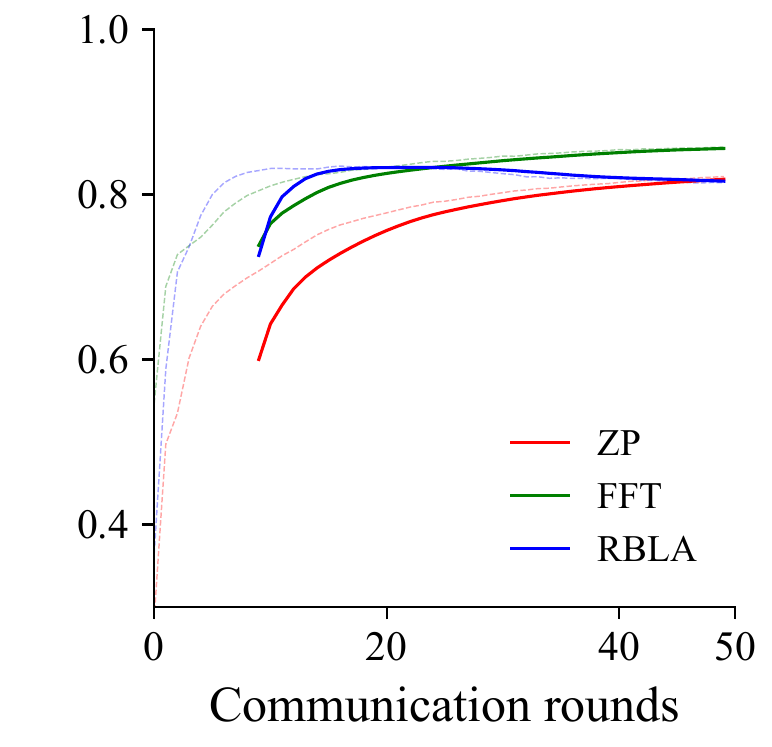}
\vspace{-1mm}
\caption{Full participation}
\label{fig:fmnist_all_participate_2nn}
\end{subfigure}
\begin{subfigure}{0.4\textwidth}
\includegraphics[width=0.9\linewidth]{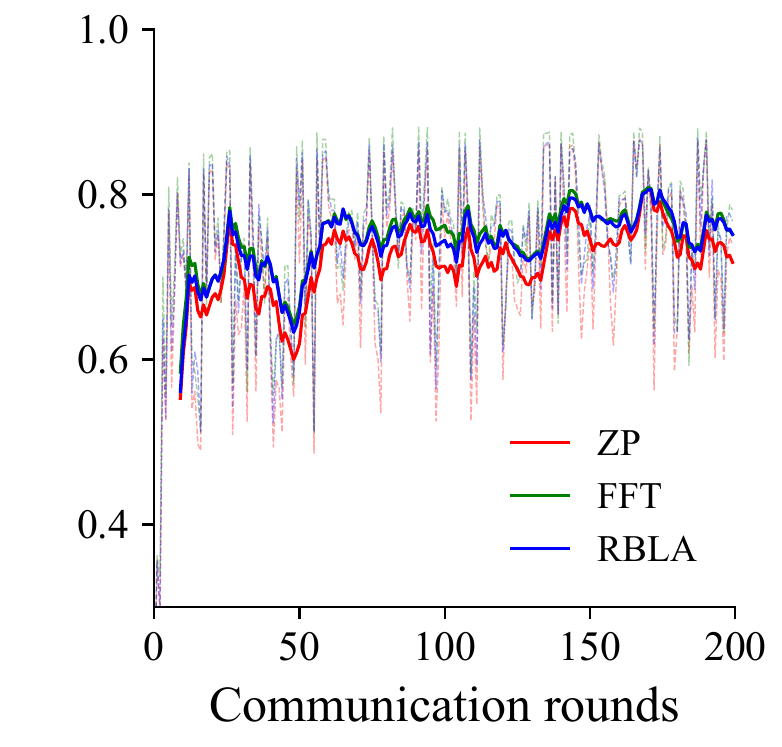}
\vspace{-1mm}
\caption{Random selection}
\label{fig:fmnist_random_selection_2nn}
\end{subfigure}
\vspace{-1mm}
\caption{
Evaluation of RBLA for \textbf{FMNIST} dataset with MLP model.}
\label{fig:fmnist_2nn}
\end{figure}

\begin{figure}[t]
\centering
\begin{subfigure}{0.4\textwidth}
\includegraphics[width=0.9\linewidth]{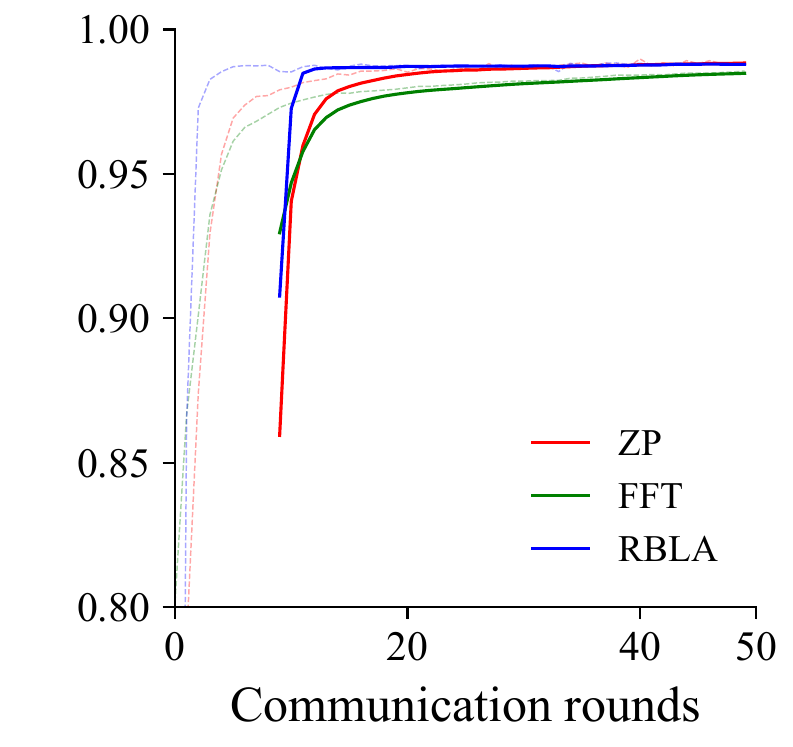} 
\vspace{-1mm}
\caption{Full participate}
\label{fig:mnist_all_participate_cnn}
\end{subfigure}
\begin{subfigure}{0.4\textwidth}
\includegraphics[width=0.9\linewidth]{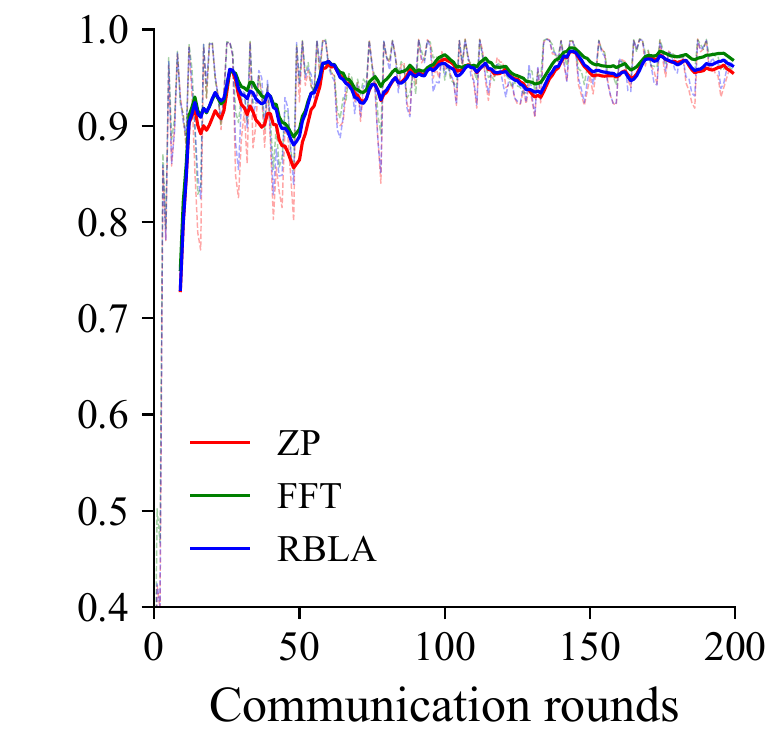}
\vspace{-1mm}
\caption{Random selection}
\label{fig:mnist_random_selection_cnn}
\end{subfigure}
\vspace{-1mm}
\caption{
Evaluation of RBLA for \textbf{MNIST} dataset with CNN model.}
\label{fig:mnist_cnn}
\end{figure}
\begin{figure}[t]
\centering
\begin{subfigure}{0.4\textwidth}
\includegraphics[width=0.9\linewidth]{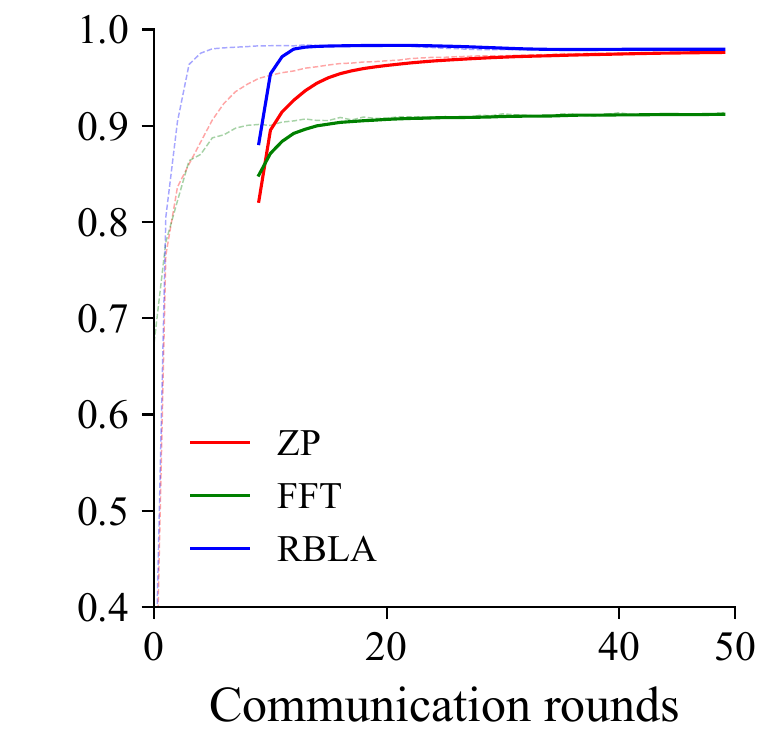}
\vspace{-1mm}
\caption{Full participation}
\label{fig:fmnist_all_participate_cnn}
\end{subfigure}
\begin{subfigure}{0.4\textwidth}
\includegraphics[width=0.9\linewidth]{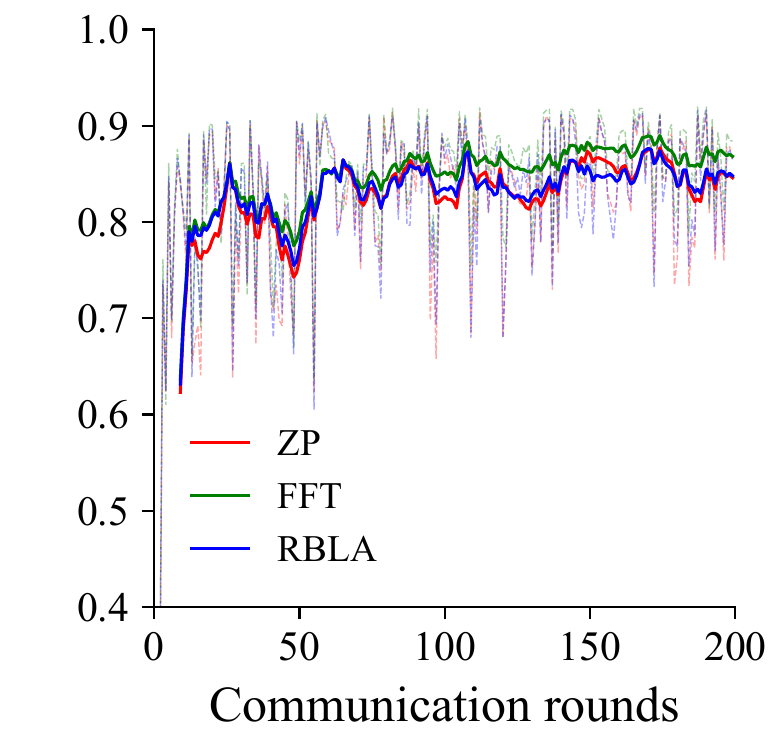}
\vspace{-1mm}
\caption{Random selection}
\label{fig:fmnist_random_selection_cnn}
\end{subfigure}
\vspace{-1mm}
\caption{
Evaluation of RBLA for \textbf{FMNIST} dataset with CNN model.}
\label{fig:fmnist_cnn}
\end{figure}

In the MNIST MLP experiments, RBLA demonstrates a superior convergence rate under the full participation setting, reaching a test accuracy of \textbf{95\%} at the \textbf{11th} communication round. In contrast, zero-padding and FFT require \textbf{42} and \textbf{40 rounds}, respectively, to reach similar accuracy levels. This indicates that RBLA reduces the required training rounds by \textbf{40\%} compared to zero-padding. In the random selection scenario, RBLA maintains a higher accuracy with less fluctuation throughout the training process compared to zero-padding and achieves performance close to FFT, which exhibits better variability.

\begin{figure}[t]
\centering
\begin{subfigure}{0.4\textwidth}
\includegraphics[width=0.9\linewidth]{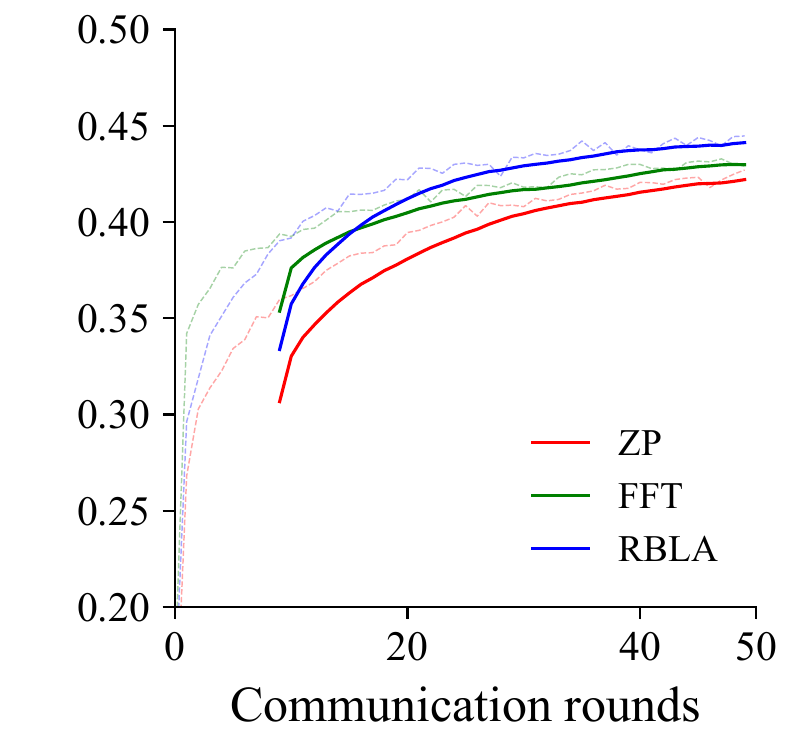}
\vspace{-1mm}
\caption{Full participation}
\label{fig:cinic_all_participate_cnn}
\end{subfigure}
\begin{subfigure}{0.4\textwidth}
\includegraphics[width=0.9\linewidth]{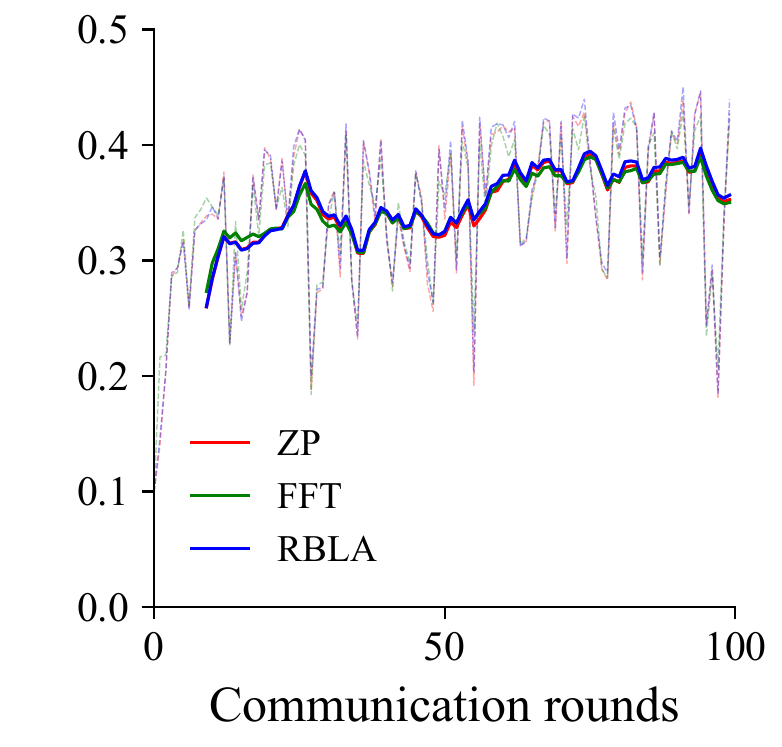}
\vspace{-1mm}
\caption{Random selection}
\label{fig:cinic_random_selection_cnn}
\end{subfigure}
\vspace{-1mm}
\caption{
Evaluation of RBLA for \textbf{CINIC-10} dataset with CNN model.}
\label{fig:cinic_cnn}
\end{figure}

Figure. \ref{fig:fmnist_2nn} shows the FMNIST MLP experiments, where RBLA again outperforms the other methods, reaching \textbf{80\%} test accuracy by the \textbf{7th} communication round, whereas FFT achieves the same accuracy after \textbf{4 rounds}, and zero-padding fails to reach this target accuracy. This demonstrates that RBLA can reduce the number of training rounds by \textbf{36\%} compared to FFT and performs significantly better than zero-padding. Figures. \ref{fig:mnist_cnn} and \ref{fig:fmnist_cnn} show the experimental results on CNN models using MNIST and FMNIST, respectively. In both cases, the advantage of RBLA becomes even more evident. In the MNIST full participation setting, RBLA achieves \textbf{98\%} test accuracy by the \textbf{4th} communication round, which is faster than FFT by \textbf{18 rounds} and faster than zero-padding by \textbf{7 rounds}. Under the random selection setting, RBLA's learning curve closely matches that of FFT and outperforms zero-padding.
In the FMNIST CNN experiments, as shown in Figure. \ref{fig:fmnist_cnn}, RBLA shows its efficiency by achieving \textbf{98\%} test accuracy of the global model at the \textbf{7th} communication round, outperforming zero-padding, which only reaches \textbf{97\%} test accuracy at the \textbf{24th} communication round, while FFT fails to converge to the target accuracy. These results consistently demonstrate that RBLA converges faster and maintains higher stability across different datasets and model types. The significant reduction in training rounds and the lower fluctuation in test accuracy further confirm the effectiveness of RBLA over traditional zero-padding methods, particularly in non-IID settings. Finally, the CIFAR-10 and CINIC-10 experiments are shown in Figures \ref{fig:cifar_cnn} and \ref{fig:cinic_cnn}, respectively. RBLA gradually closes the performance gap with FFT as training rounds increase, outperforming zero-padding. While RBLA and FFT show similar convergence speeds at the beginning of training, RBLA converges more slowly after several rounds in the CIFAR-10 experiment but exhibits a faster convergence rate than both zero-padding and FFT in the CINIC-10 experiments.
\vspace{-5mm}
\section{Conclusion}

This paper proposes RBLA, a model aggregation method specifically designed for heterogeneous LoRA models in Federated Learning as a Service (FLaaS) systems. RBLA addresses the critical challenge of preserving both low-rank and high-rank features during the aggregation process, which becomes particularly complex in scenarios with diverse client models. By leveraging FLaaS, RBLA enhances the convergence rate of the global model in non-IID scenarios with heterogeneous model structures. This improvement is effective in real-world applications with varied client device performance and skewed data distributions. The experimental results demonstrate RBLA's practicality and efficiency in practical FLaaS scenarios.
\vspace{-3mm}
\bibliography{citation.bib}
\bibliographystyle{splncs04}
\end{document}